\documentclass[11pt]{article}
\usepackage{coling2018}
\usepackage{times}
\usepackage{url}
\usepackage{latexsym}

\usepackage{multirow}
\usepackage{array}
\usepackage{url}
\usepackage{caption}
\usepackage{chngpage}
\usepackage{subfigure}
\usepackage{graphicx}
\usepackage{amsfonts}
\usepackage{amsmath}{}
\usepackage{CJK}
\usepackage{multirow}
\usepackage{enumerate}
\usepackage{xcolor}
\newcommand{\ignore}[1]{} 

\title{Modeling Coherence for Neural Machine Translation \\ with Dynamic and Topic Caches}

\author{Shaohui Kuang$^\dagger$\hspace{1cm} Deyi Xiong$^\dagger$\thanks{ \hspace{0.1cm} Corresponding author}\hspace{1cm}  Weihua Luo$^{\dagger\dagger}$\hspace{1cm} Guodong Zhou$^\dagger$ \\
       $^\dagger$School of Computer Science and Technology, Soochow University, Suzhou, China\\ 
       {\tt shkuang@stu.suda.edu.cn, \{dyxiong, gdzhou\}@suda.edu.cn} \\
       $^{\dagger\dagger}$Alibaba Group \\
       {\tt weihua.luowh@alibaba-inc.com}\\}

\date{}

\begin{document}
\begin{CJK}{UTF8}{gbsn}
\maketitle
\begin{abstract}

Sentences in a well-formed text are connected to each other via various links to form the cohesive structure of the text. Current neural machine translation (NMT) systems translate a text in a conventional sentence-by-sentence fashion, ignoring such cross-sentence links and dependencies. This may lead to generate an incoherent target text for a coherent source text. In order to handle this issue, we propose a cache-based approach to modeling coherence for neural machine translation by capturing contextual information either from recently translated sentences or the entire document. Particularly, we explore two types of caches:  a dynamic cache, which stores words from the best translation hypotheses of preceding sentences, and a topic cache, which maintains a set of  target-side topical words that are semantically related to the document to be translated. On this basis, we build a new layer to score target words in these two caches with a cache-based neural model. Here the estimated probabilities from the cache-based neural model are combined with NMT probabilities into the final word prediction probabilities via a gating mechanism. Finally, the proposed cache-based neural model is trained jointly with NMT system in an end-to-end manner. 
Experiments and analysis presented in this paper demonstrate that the proposed cache-based model achieves substantial improvements over several state-of-the-art SMT and NMT baselines.

\end{abstract}


    %
    %
    
    %
    %
    %
    %

Neural machine translation \cite{sutskever2014sequence,bahdanau2015neural} as an emerging machine translation approach, quickly achieves the state-of-the-art translation performance on many language pairs, e.g., English-French \cite{Jean2014On,luong2014addressing}, English-German \cite{Shen2015Minimum,luong2015effective} and so on. In principle, NMT is established on an encoder-decoder framework, where the encoder reads a source sentence and encodes it into a fixed-length semantic vector, and the decoder generates a translation according to this vector. 

In spite of its current success, NMT translates sentences of a text independently, ignoring document-level information during translation. This largely limits its success since document-level information imposes constraints on the translations of individual sentences of a text. And such document-level constraints, at least, can be categorized into three aspects: consistency, disambiguation and coherence.  First, the same phrases or terms should be translated consistently across the entire text as much as possible, no matter in which sentence they occur.  If sentences of a text are translated independent of each other, it will be difficult to maintain the translation consistency across different sentences. Second, document-level information provides a global context that can help disambiguate words with multiple senses if sentence-level local context is not sufficient for disambiguation. Third, the topic of a document is able to keep individual sentences translated in a coherent way. 

In the literature, such informative constraints have been occasionally investigated in statistical machine translation and achieved certain success via a variety of document-level models, such as cache-based language and translation models \cite{tiedemann2010context,gong2011cache,nepveu2004adaptive} for the consistency constraint, topic-based coherence model \cite{xiong2013topic,Tam2007Bilingual} for the coherence constraint. By contrast, in neural machine translation, to the best of our knowledge, such constraints have not been explored so far. 

Partially inspired by the success of cache models in SMT, we propose a cache-based approach to capturing coherence for neural machine translation. Particularly, we incorporate two types of caches into NMT: a static topic cache and a dynamic cache being updated on the fly. For the topic cache, we first use a projection-based bilingual topic learning approach to infer the topic distribution for each document to be translated and obtain the corresponding topical words on the target side. These topical words are then integrated into the topic cache.  For the dynamic cache, words are retrieved from the best translation hypotheses of recently translated sentences. While the topic cache builds a global profile for a document, which helps impose the coherence constraint on document translation, the dynamic cache follows an intuition that words occurred previously should have higher probabilities of recurrence even if they are rare words in the training data.  

In order to integrate these two caches into neural machine translation, we further propose a cache-based neural model, which adds a new neural network layer on the cache component to score each word in the cache. During decoding, we estimate the probability of a word from the cache according to its score and combine this cache probability with the original probability computed by the decoder via a gating mechanism to obtain the final word prediction probability. 

On the NIST Chinese-English translation tasks, our experiment results show that the proposed cache-based neural approach can achieve significant improvements of up to 1.60 BLEU points on average (up to 2.53 BLEU points on NIST04) over the state-of-the-art attention-based NMT baseline.


\section{Related Work}

In the literature, several cache-based translation models have been proposed for conventional statistical machine translation, besides traditional n-gram language models and neural language models. In this section, we will first introduce related work in cache-based language models and then in translation models.

For traditional n-gram language models, 
\newcite{Kuhn1990A} propose a cache-based language model, which mixes a large global language model with a small local model estimated from recent items in the history of the input stream for speech recongnition. 
\newcite{della1992adaptive} introduce a MaxEnt-based cache model by integrating a cache into a smoothed trigram language model, reporting reduction in both perplexity and word error rates. 
\newcite{chueh2010topic} present a new topic cache model for speech recongnition based on latent Dirichlet language model by incorporating a large-span topic cache into the generation of topic mixtures. 

For neural language models,
\newcite{huang2014cache} propose a cache-based RNN inference scheme, which avoids repeated computation of identical LM calls and caches previously computed scores and useful intermediate results and thus reduce the computational expense of RNNLM. 
\newcite{Grave2016Improving} extend the neural network language model with a neural cache model, which stores recent hidden activations to be used as contextual representations. 
Our caches significantly differ from these two caches in that we store linguistic items in the cache rather than scores or activations.

For neural machine translation, \newcite{wangexploiting}  propose a cross-sentence context-aware approach and employ a hierarchy of Recurrent Neural Networks (RNNs) to summarize the cross-sentence context from source-side previous sentences.
\newcite{jean2017does} propose a novel larger-context neural machine translation model based on the recent works on larger-context language modelling \cite{wang2016larger} and employ the method to model the surrounding text in addition to the source sentence.

For cache-based translation models, 
\newcite{nepveu2004adaptive} propose a dynamic adaptive translation model using cache-based implementation for interactive machine translation, and develop a monolingual dynamic adaptive model and a bilingual dynamic adaptive model.  
\newcite{tiedemann2010context} propose a cache-based translation model, filling the cache with bilingual phrase pairs from the best translation hypotheses of previous sentences in a document.  
\newcite{gong2011cache} further propose a cache-based approach to document-level translation, which includes three caches, a dynamic cache, a static cache and a topic cache, to capture various document-level information.  
\newcite{bertoldi2013cache} describe a cache mechanism to implement online learning in phrase-based SMT and use a repetition rate measure to predict the utility of cached items expected to be useful for the current translation.

Our caches are similar to those used by \newcite{gong2011cache} who incorporate these caches into statistical machine translation. We adapt them to neural machine translation with a neural cache model. It is worthwhile to emphasize that such adaptation is nontrivial as shown below because the two translation philosophies and frameworks are significantly different.

\section{Attention-based NMT}
In this section, we briefly describe the NMT model taken as a baseline. Without loss of generality, we adopt the NMT architecture proposed by \newcite{bahdanau2015neural}, with an encoder-decoder neural network. 

\subsection{Encoder} 

The encoder uses bidirectional recurrent neural networks (Bi-RNN) to encode a source sentence with a forward and a backward RNN. The forward RNN takes as input a source sentence \(x = (x_1, x_2, ..., x_T)\) from left to right and outputs a hidden state sequence \((\overrightarrow{h_1},\overrightarrow{h_2}, ..., \overrightarrow{h_T})\) while the backward RNN reads the sentence in an inverse direction and outputs a backward hidden state sequence \((\overleftarrow{h_1},\overleftarrow{h_2}, ..., \overleftarrow{h_T})\). The context-dependent word representations of the source sentence \(h_j\) (also known as word annotation vectors) are the concatenation of hidden states \(\overrightarrow{h_j}\) and \(\overleftarrow{h_j}\) in the two directions.

\subsection{Decoder}

The decoder is an RNN that predicts target words \(y_t\)  via a multi-layer perceptron (MLP) neural network. The prediction is based on the decoder RNN hidden state \(s_t\), the previous predicted word \(y_{t-1}\) and a source-side context vector \(c_t\). The hidden state \(s_t\) of the decoder at time \(t\) and the conditional probability of the next word \(y_t\) are computed as follows:

\begin{equation}
s_t = f(s_{t-1}, y_{t-1}, c_t)
\end{equation}

\begin{equation}
p(y_t|y_{<t};x) = g(y_{t-1}, s_t, c_t)
\end{equation}

\subsection{Attention Model}

In the attention model, the context vector \(c_t\) is calculated as a weighted sum over source annotation vectors \((h_1, h_2, ..., h_T)\):
\begin{equation}
c_t = \sum_{j=1}^{T_x} \alpha_{tj}h_j
\end{equation}
\begin{equation}
\alpha_{tj} = \frac{exp(e_{tj})}{\sum_{k=1}^{T} exp(e_{tk})}
\end{equation}
\begin{equation}
e_{tj} = a(s_{t-1},h_j)
\end{equation}
where \(\alpha_{tj}\) is the attention weight of each hidden state \(h_j\) computed by the attention model, and \(a\) is a feed forward neural network with a single hidden layer.

The dl4mt tutorial\footnote{\url{https://github.com/nyu-dl/dl4mt-tutorial/tree/master/session2}} presents an improved implementation of the attention-based NMT system, which feeds the previous word \(y_{t-1}\) to the attention model. We use the dl4mt tutorial  implementation as our baseline, which we will refer to as RNNSearch*.


The proposed cache-based neural approach is implemented on the top of RNNSearch* system, where the encoder-decoder NMT framework is trained to optimize the sum of the conditional log probabilities of correct translations of all source sentences on a parallel corpus as normal. 

\section{The Cache-based Neural Model}

In this section, we elaborate on the proposed cache-based neural model and how we integrate it into neural machine translation,  Figure 1 shows the entire architecture of our NMT with the cache-based neural model. 

\begin{figure*}[!t]
\centering
\includegraphics[height=1.2in,width=3.5in]{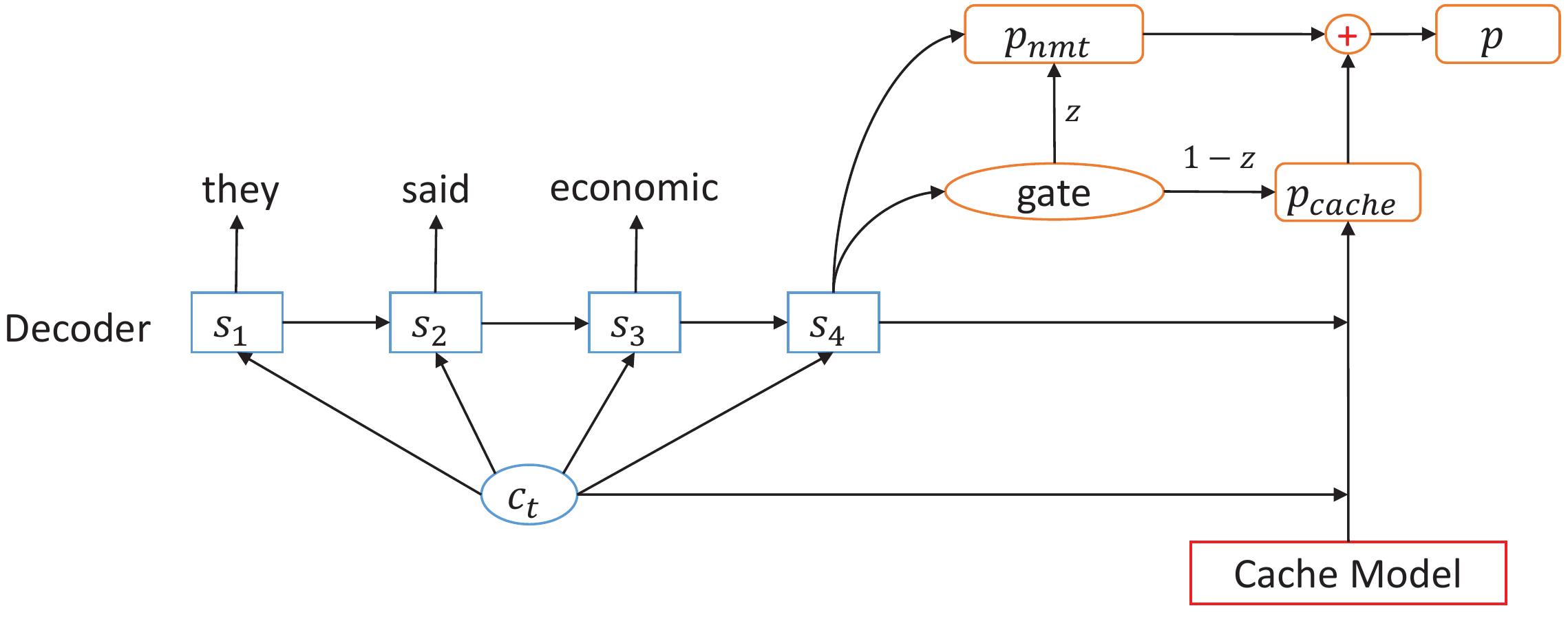}
\caption{Architecture of NMT with the neural cache model. \(P_{cache}\) is the probability for a next target word estimated by the cache-based neural model.}
\label{fig:1}
\end{figure*}

\subsection{Dynamic Cache and Topic Cache}

The aim of cache is to incorporate document-level constraints and therefore to improve the consistency and coherence of document translations. In this section, we introduce our proposed dynamic cache and topic cache in detail.

\subsubsection{Dynamic Cache}
In order to build the dynamic cache, we dynamically extract words from recently translated sentences and the partial translation of current sentence being translated as words of dynamic cache. We apply the following rules to build the dynamic cache.
\begin{enumerate}[a)]
\item The max size of the dynamic cache is set to \(|c_d|\). 
\item According to the first-in-first-out rule, when the dynamic cache is full and a new word is inserted into the cache, the oldest word in the cache will be removed. 
\item Duplicate entries into the dynamic cache are not allowed when a word has been already in the cache.
\end{enumerate}

It is worth noting that we also maintain a stop word list, and we added English punctuations and ``UNK'' into our stop word list. Words in the stop word list would not be inserted into the dynamic cache. So the common words like ``a'' and ``the'' cannot appear in the cache. 
All words in the dynamic cache can be found in the target-side vocabulary of RNNSearch*. 

\subsubsection{Topic Cache}
In order to build the topic cache, we first use an off-the-shelf LDA topic tool\footnote{http://www.arbylon.net/projects/} to learn topic distributions of source- and target-side documents separately. Then we estimate a topic projection distribution over all target-side topics \(p(z_t|z_s)\) for each source topic \(z_s\) by collecting events and accumulating counts of \((z_s, z_t)\) from aligned document pairs. Notice that \(z_s/z_t\) is the topic with the highest topic probability \(p(z_.|d)\) on the source/target side. Then we can use the topic cache as follows:

\begin{enumerate}[1)]
\item During the training process of NMT, the learned target-side topic model is used to infer the topic distribution for each target document. For a target document d in the training data, we select the topic \(z\) with the highest probability \(p(z|d)\) as the topic for the document. The \(|c_t|\) most probable topical words in topic \(z\) are extracted to fill the topic cache for the document \(d\).

\item In the NMT testing process, we first infer the topic distribution for a source document in question with the learned source-side topic model. From the topic distribution, we choose the topic with the highest probability as the topic for the source document. Then we use the learned topic projection function to map the source topic onto a target topic with the highest projection probability, as illustrated in Figure 2. After that, we use the \(|c_t|\) most probable topical words in the projected target topic to fill the topic cache. 
\end{enumerate}

The words of topic cache and dynamic cache together form the final cache model. 
In practice, the cache stores word embeddings, as shown in Figure 3. As we do not want to introduce extra embedding
parameters, we let the cache share the same target word embedding matrix with the NMT model. In this case, if a word is not in the target-side vocabulary of NMT, we discard the word from the cache.

\subsection{The Model}
The cache-based neural model is to evaluate the probabilities of words occurring in the cache and to provide the evaluation results for the decoder via a gating mechanism. 

\subsubsection{Evaluating Word Entries in the Cache}

When the decoder generates the next target word \(y_t\), we hope that the cache can provide helpful information to judge whether \(y_t\) is appropriate from the perspective of the document-level cache if \(y_t\) occurs in the cache.To achieve this goal, we should appropriately evaluate the word entries in the cache. 

In this paper, we build a new neural network layer as the scorer for the cache. At each decoding step \(t\), we use the scorer to score \(y_t\) if \(y_t\) is in the cache. The inputs to the scorer are the current hidden state \(s_t\) of the decoder, previous word \(y_{t-1}\), context vector \(c_t\), and the word \(y_t\) from the cache. The score of \(y_t\) is calculated as follows:
\begin{equation}
score(y_t|y_{<t},x) = g_{cache}(s_t,c_t,y_{t-1},y_t)
\end{equation}
where \(g_{cache}\) is a non-linear function.

This score is further used to estimate the cache probability of \(y_t\) as follows:
\begin{equation}
p_{cache}(y_t|y_{<t},x) = softmax(score(y_t|y_{<t},x))
\end{equation}

\begin{figure}  
\begin{minipage}[t]{0.5\linewidth}  
\centering  
\includegraphics[height=1.3in,width=1.2in]{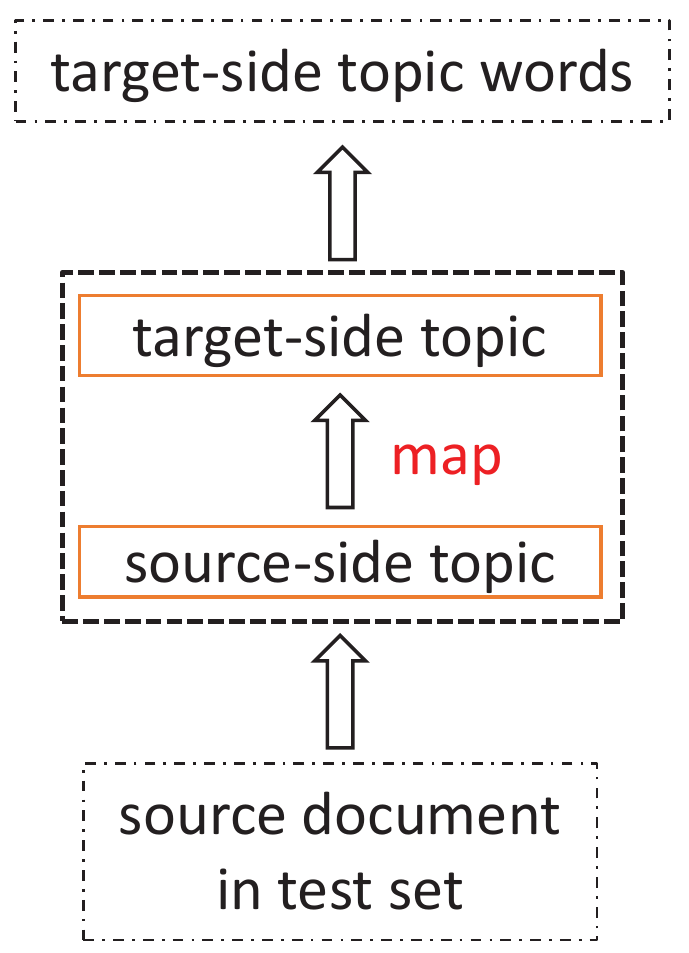}  
\caption{Schematic diagram of the topic projection during the testing process.}  
\label{fig:side:a}  
\end{minipage}%
\begin{minipage}[t]{0.5\linewidth}  
\centering  
\includegraphics[height=1.3in,width=2.2in]{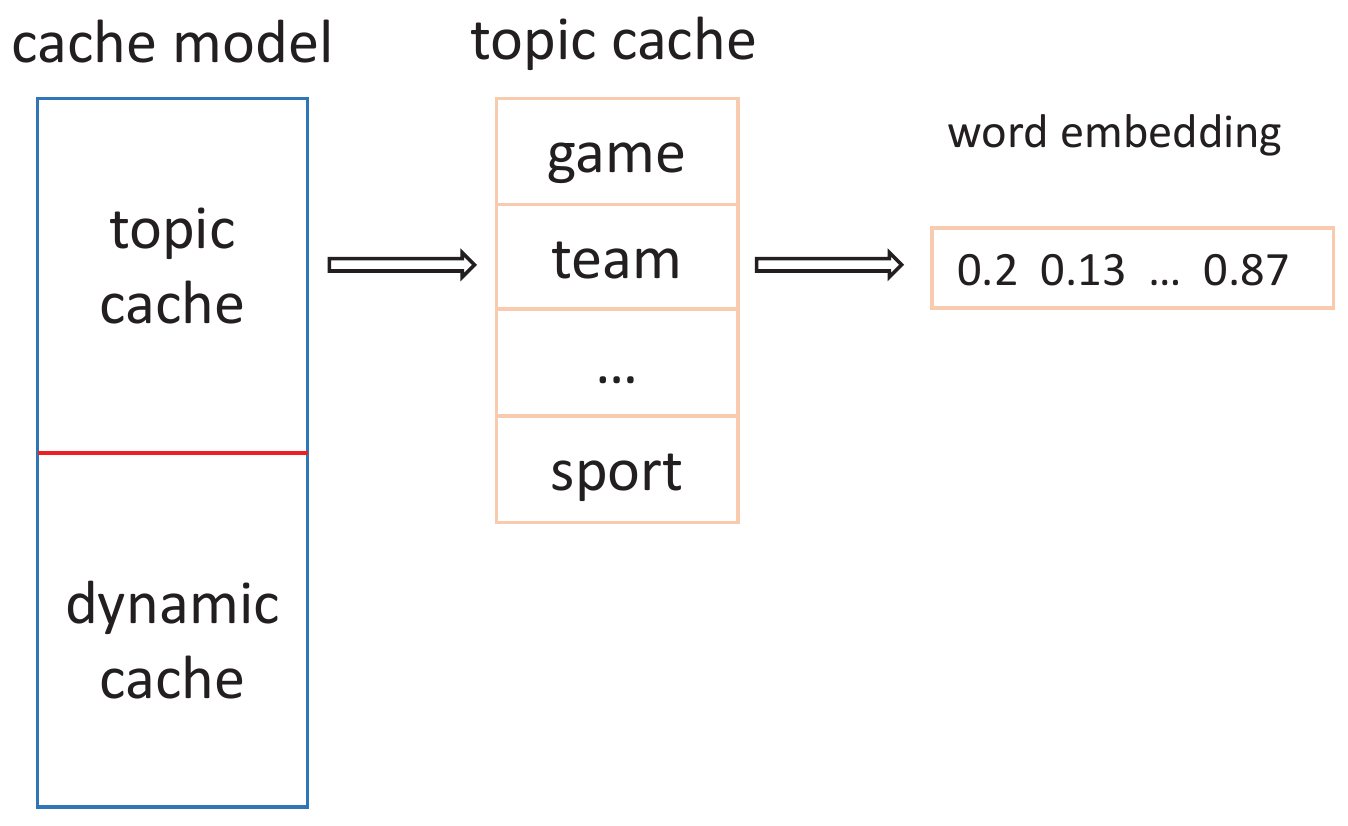}  
\caption{Architecture of the cache model.}  
\label{fig:side:b}  
\end{minipage}  
\end{figure}

\subsubsection{Integrating the Cache-based Neural Model into NMT}

Since we have two prediction probabilities for the next target word \(y_t\), one from the cache-based neural model \(p_{cache}\), the other originally estimated by the NMT decoder \(p_{nmt}\), how do we integrate these two probabilities?  Here, we introduce a gating  mechanism  to combine them, and word prediction probabilities on the vocabulary of NMT are updated by combining the two probabilities through linear interpolation between the NMT probability and cache-based neural model probability. The final word prediction probability for \(y_t\) is calculated as follows:
\begin{equation}
p(y_t|y_{<t},x) = (1 - \alpha_t)p_{cache}(y_t|y_{<t},x)  + \alpha_tp_{nmt}(y_t|y_{<t},x) 
\end{equation}
Notice that if \(y_t\) is not in the cache, we set \(p_{cache}(y_t|y_{<t},x) = 0\), where \(\alpha_t\) is the gate and computed as follows:
\begin{equation}
  \alpha_t = g_{gate}(f_{gate}(s_t,c_t,y_{t-1}))
\end{equation}
where \(f_{gate}\) is a non-linear function and $g_{gate}$ is sigmoid function.


We use the contextual elements of $s_t, c_t, y_{t-1}$ to score the current target word occurring in the cache (Eq. (6)) and to estimate the gate (Eq. (9)). If the target word is consistent with the context and in the cache at the same time, the probability of the target word will be high.

Finally, we train the proposed cache model jointly with the NMT model towards minimizing the negative log-likelihood on the training corpus. The cost function is computed as follows:
\begin{equation}
L(\theta) = -\sum_{i=1}^N \sum_{t=1}^Tlogp(y_t|y_{<t},x)
\end{equation}
where \(\theta\) are all parameters in the cache-based NMT model.

\subsection{Decoding Process}

Our cache-based NMT system works as follows:
\begin{enumerate}[(1)]
\item When the decoder shifts to a new test document, clear the topic and dynamic cache.
\item Obtain target topical words for the new test document as described in Section 4.1 and fill them in the topic cache.  
\item Clear the dynamic cache when translating the first sentence of the test document. 
\item For each sentence in the new test document, translate it with the proposed cache-based NMT and continuously expands the dynamic cache with newly generated target words and target words obtained from the best translation hypothesis of previous sentences. 
\end{enumerate}

In this way, the topic cache can provide useful global information at the beginning of the translation process while the dynamic cache is growing with the progress of translation.

\section{Experimentation}

We evaluated the effectiveness of the proposed cache-based neural model for neural machine translation on NIST Chinese-English translation tasks. 

\subsection{Experimental Setting}

We selected corpora LDC2003E14, LDC2004T07, LDC2005T06, LDC2005T10 and a portion of data from the corpus LDC2004T08 (Hong Kong Hansards/Laws/News) as our bilingual training data, where document boundaries are explicitly kept. In total, our training data contain 103,236 documents and 2.80M sentences. On average, each document consists of 28.4 sentences. We chose NIST05 dataset (1082 sentence pairs) as our development set, and NIST02, NIST04, NIST06 (878, 1788, 1664 sentence pairs. respectively) as our test sets.  We compared our proposed model against the following two systems:
\begin{itemize}
\item {\bf Moses} \cite{koehn2007moses}: an off-the-shelf phrase-based translation system with its default setting.
\item {\bf RNNSearch*}: our in-house attention-based NMT system which adopts the feedback attention as described in Section 3
. 
\end{itemize}

For Moses, we used the full training data to train the model. We ran GIZA++ \cite{och2000improved} on the training data in both directions, and merged alignments in two directions with  ``grow-diag-final'' refinement rule \cite{koehn2005edinburgh} to obtain final word alignments. We trained a 5-gram language model on the Xinhua portion of GIGA-WORD corpus using SRILM Toolkit with a modified Kneser-Ney smoothing.

For RNNSearch, we used the parallel corpus to train the attention-based NMT model. The encoder of RNNSearch consists of a forward and backward recurrent neural network. The word embedding dimension is 620 and the size of a hidden layer is 1000.  
The maximum length of sentences that we used to train RNNSearch in our experiments was set to 50 on both Chinese and English side. We used the most frequent 30K words for both Chinese and English. We replaced rare words with a special token ``UNK''. Dropout was applied only on the output layer and the dropout rate was set to 0.5. All the other settings were the same as those in \cite{bahdanau2015neural}. Once the NMT model was trained, we adopted a beam search to find possible translations with high probabilities. We set the beam width to 10.

For the proposed cache-based NMT model, we implemented it on the top of RNNSearch*. 
We set the size of the dynamic and topic cache \(|c_d|\) and \(|c_t|\) to 100, 200, respectively. For the dynamic cache, we only kept those most recently-visited items.  For the LDA tool, we set the number of topics considered in the model to 100 and set the number of topic words that are used to fill the topic cache to 200. The parameter \(\alpha\) and \(\beta\) of LDA were set to 0.5 and 0.1, respectively.
We used a feedforward neural network with two hidden layers to define $g_{cache}$ (Equation (6)) and $f_{gate}$ (Equation (9)). For $f_{gate}$, the number of units in the two hidden layers were set to 500 and 200 respectively. For $g_{cache}$, the number of units in the two hidden layers were set to 1000 and 500 respectively.
We used a pre-training strategy that has been widely used in the literature to train our proposed model: 
training the regular attention-based NMT model using our implementation of RNNSearch*, and then using its parameters to initialize the parameters of the proposed model, except for those related to the operations of the proposed cache model. 

We used the stochastic gradient descent algorithm with mini-batch and Adadelta to train the NMT models. The mini-batch was set to 80 sentences and decay rates $\rho$ and $\epsilon$ of Adadelta were set to 0.95 and \(10^{-6}\).

\subsection{Experimental Results}

Table 1 shows the results of different models measured in terms of BLEU score\footnote{As our model requires document boundaries so as to gurantee that cache words are from the same document, we use all LDC corpora that provide document boundaries. Most training sentences are from Hong Kong Hansards/Laws Parallel Text (accounting for 57.82\% of our training data) which are in the law domain rather than the news domain of our test/dev sets. This is the reason that our baseline is lower than other published results obtained using more news-domain training data without document boundaries.}. 
From the table, we can find that our implementation RNNSearch* using the feedback attention and dropout outperforms Moses by 3.23 BLEU points. The proposed model \(RNNSearch*_{+Cd}\) achieves an average gain of 1.01 BLEU points over RNNSearch* on all test sets. Further, the model \(RNNSearch*_{+Cd, Ct}\) achieves an average gain of 1.60 BLEU points over RNNSearch*, and it outperforms Moses by 4.83 BLEU points.  These results strongly suggest that the dynamic and topic cache are very helpful and able to improve translation quality in document translation. 

\begin{table*}[!t]
\centering
\small
\begin{tabular}{lccccccc}
\hline 
\bf Model & \bf NIST02 & \bf NIST04 & \bf NIST05 & \bf NIST06 & \bf Avg\\ 
\hline
Moses              & 31.52  & 32.73 & 29.52 & 29.57  & 30.69 \\
\hline
RNNSearch*          & 36.18 & 36.36 &32.56 & 30.57 & 33.92\\
+ \(Cd\)           & 36.86 & 37.16 & 33.10 & 32.60  & 34.93\\
+ \(Cd, Ct\)       & 38.04 & 38.89 & 33.31 & 31.85 & 35.52 \\
\hline
\end{tabular}
\caption{\label{font-table} Experiment results on the NIST Chinese-English translation tasks. [+\(Cd\)] is the proposed model with the dynamic cache. [+\(Cd, Ct\)] is the proposed model with both the dynamic and topic cache. 
 Avg means the average BLEU score on all test sets.}
\end{table*}

\subsection*{Effect of the Gating Mechanism}

In order to validate the effectiveness of the gating mechanism used in the cache-based neural model, we set a fixed gate value for \(RNNSearch*_{+Cd,Ct}\), in other words, we use a mixture of probabilities with fixed proportions to replace the gating mechanism that automatically learns weights for probability mixture. 

Table 2 displays the result. 
When we set the gate \(\alpha\) to a fixed value 0.3,  the performance has an obvious decline comparing with that of \(RNNSearch*_{+Cd,Ct}\) in terms of BLEU score. The performance is even worse than RNNSearch* by 10.11 BLEU points. Therefore without a good mechanism, the cache-based neural model cannot be appropriately integrated into NMT. This shows that the gating mechanism plays a important role in \(RNNSearch*_{+Cd,Ct}\).

\begin{table*}[!t]
\centering
\small
\begin{tabular}{c|ccccccc}
\hline 
\bf Model & \bf NIST02 & \bf NIST04 & \bf NIST05 & \bf NIST06 & \bf Avg\\ 
\hline
RNNSearch*          & 36.18 & 36.36 &32.56 & 30.57 & 33.92 \\
\hline
+ \(Cd, Ct\)       & 38.04 & 38.89 & 33.31 & 31.85 & 35.52 \\
+ \(\alpha = 0.3\) & 23.39  & 17.83 & 31.51 & 28.90 & 25.41 \\ 
\hline
\end{tabular}
\caption{\label{font-table2} Effect of the gating mechanism. 
[+$\alpha$=0.3] is the [\(+Cd, Ct\)] with a fixed gate value 0.3.}
\end{table*}

\subsection*{Effect of the Topic Cache}

When the NMT decoder translates the first sentence of a document, the dynamic cache is empty. In this case, we hope that the topic cache will provide document-level information for translating the first sentence. We therefore further investigate how the topic cache influence the translation of the first sentence in a document. We count and calculate the average number of words that appear in both the translations of the beginning sentences of documents and the topic cache.

The statistical results are shown in Table 3. 
Without using the cache model, RNNSearch* generates translations that contain words from the topic cache as these topic words are tightly related to documents being translated.  
With the topic cache, our neural cache model enables the translations of the first sentences to be more relevant to the global topics of documents being translated as these translations contain more words from the topic cache that describes these documents. As the dynamic cache is empty when the decoder translates the beginning sentences, the topic cache is complementary to such a cold cache at the start. Comparing the numbers of translations generated by our model and human translations (Reference in Table 3), we can find that with the help of the topic cache, translations of the first sentences of documents are becoming closer to human translations.

\begin{table*}[!t]
\centering
\small
\begin{tabular}{l|ccccccc}
\hline 
\bf Model & \bf NIST02 & \bf NIST04 & \bf NIST05 & \bf NIST06  & \bf Mean \\ 
\hline
RNNSearch*          & 1.90  & 2.43 & 1.31 & 1.50 & 1.78  \\
+ \(Ct\)            & 2.11  & 2.51 & 1.53 & 1.73 & 1.96  \\
Reference           & 2.39  & 2.51 & 2.20 & 1.28 & 2.09  \\
\hline
\end{tabular}
\caption{\label{font-table3} The average number of words in translations of beginning sentences of documents that are also in the topic cache. Reference represents the average number of words in four human translations that are also in the topic cache.}
\end{table*}

\subsection*{Analysis on the Cache-based Neural Model}

As shown above, the topic cache is able to influence both the translations of beginning sentences and those of subsequent sentences while the dynamic cache built from translations of preceding sentences has an impact on the translations of subsequent sentences. We further study what roles the dynamic and topic cache play in the translation process. For this aim, we calculate the average number of words in translations generated by \(RNNSearch*_{+Cd,Ct}\) that are also in the caches.
During the counting process, stop words and ``UNK'' are removed from sentence and document translations. 

Table 4 shows the results. 
If only the topic cache is used ([\(document \in [Ct]\),\(sentence \in (Ct)\)] in Table 4), the cache still can provide useful information to help NMT translate sentences and documents. 28.3 words per document and 2.39 words per sentence are from the topic cache. 
When both the dynamic and topic cache are used ([\(document \in [Ct,Cd]\),\(sentence \in (Ct,Cd)\)] in Table 4), the numbers of words that both occur in sentence/document translations and the two caches sharply increase from 2.61/30.27 to 6.73/81.16.  
The reason for this is that words appear in preceding sentences will have a large probability of appearing in subsequent sentences. This shows that the dynamic cache plays a important role in keeping document translations consistent by reusing words from preceding sentences.

\begin{table*}[!t]
\centering
\small
\begin{tabular}{l|ccccccc}
\hline 
\bf Model & \bf NIST02 & \bf NIST04 & \bf NIST05 & \bf NIST06 &\bf Avg\\ 
\hline
\(document \in (Ct)\)      & 25.70  & 27.48 & 26.89 & 41.00  & 30.27\\
\(document \in (Cd, Ct)\)  & 59.46  & 64.24 & 76.70 & 124.25 & 81.16\\
\hline
\(sentence \in (Ct)\)      & 2.93  & 3.07 & 2.49 & 1.95 & 2.61 \\
\(sentence \in (Cd, Ct)\)  & 6.77  & 7.18 & 7.09 & 5.89 & 6.73 \\
\hline
\end{tabular}
\caption{\label{font-table4} The average number of words in translations generated by \(RNNSearch*_{+Cd,Ct}\) that are also in the dynamic and topic cache. [\(document/sentence \in [Ct]\)] denote the average number of words that are in both document/sentence translations and the topic cache. [\(document/sentence \in [Cd,Ct]\)] denote the average number of words occurring in both document/sentence translations and the two caches.}
\end{table*}

\begin{table*}[!t]
\centering
\small
\begin{tabular}{l|l}
\hline
\hline
\multirow{2}{*}{SRC} 
& (1) 并 将 计划 中 的 一 系列 军事 行动 提前 付诸 实施 。\\
& (2) 会议 决定 加大 对 巴方 的 军事 打击 行动 。 \\
\hline
\multirow{2}{*}{REF} 
& (1) and to implement ahead of schedule a series of military actions still being planned . \\
& (2) the meeting decided to increase military actions against palestinian . \\
\hline
\multirow{2}{*}{RNNSearch*} 
& (1) and to implement a series of military operations . \\
& (2) the meeting decided to increase military actions against the palestinian side . \\
\hline
\multirow{2}{*}{+ \(Cd, Ct\) } 
& (1) and to implement a series of military actions plans. \\
& (2) the meeting decided to increase military actions against the palestinian side . \\
\hline
\hline
\end{tabular}
\caption{\label{font-table5} Translation examples on the test set. SRC for source sentences, REF for human translations. These two sentences (1) and (2) are in the same document.}
\end{table*}

We also provide two translation examples in Table 5. We can find that RNNSearch* generates different translations ``operations'' and ``actions'' for the same chinese word ``行动(xingdong)'', while our proposed model produces the same translation ``actions''. 

\subsection{Analysis on Translation Coherence}

\begin{table}[!t]
\centering
\small
\begin{tabular}{c|c}
\hline
\hline
\bf Model & \bf Coherence \\
\hline
RNNSearch*            & 0.4259 \\
\(RNNSearch*_{+Cd,Ct}\) & 0.4274 \\
Human Reference         & 0.4347 \\
\hline
\hline
\end{tabular}
\caption{\label{font-table7} The average cosine similarity of adjacent sentences (coherence) on all test sets.}
\end{table}

We want to further study how the proposed cache-based neural model influence coherence in document translation. For this, we follow \newcite{Lapata2005Automatic} to measure coherence as sentence similarity. First, each sentence is represented by the mean of the distributed vectors of its words. Second, the similarity between two sentences is determined by the cosine of their means.
\begin{equation}
 sim(S_1,S_2) = cos(\mu(\vec{S_1}),\mu(\vec{S_2})) \\
\end{equation}
where \(\mu(\vec{S_i})=\frac{1}{|S_i|}\sum_{\vec{w} \in S_i}\vec{w}\), and \(\vec{w}\) is the vector for word \(w\).

We use Word2Vec\footnote{http://word2vec.googlecode.com/svn/trunk/} to get the distributed vectors of words and English Gigaword fourth Edition\footnote{https://catalog.ldc.upenn.edu/LDC2009T13} as training data to train Word2Vec. We consider that embeddings from word2vec trained on large monolingual corpus can well encode semantic information of words. We set the dimensionality of word embeddings to 200. 
Table 6 shows the average cosine similarity of adjacent sentences on all test sets. From the table, we can find that the \(RNNSearch*_{+Cd,Ct}\) model produces better coherence in document translation than RNNSearch* in term of cosine similarity.

\section{Conclusion}
In this paper, we have presented a novel cache-based neural model for NMT to capture the global topic information and inter-sentence cohesion dependencies. We use a gating mechanism to integrate both the topic and dynamic cache into the proposed neural cache model. Experiment results show that the  cache-based neural model achieves consistent and significant improvements in translation quality over several state-of-the-art NMT and SMT baselines. Further analysis reveals that the topic cache and dynamic cache are complementary to each other and that both are able to guide the NMT decoder to use topical words and to reuse words from recently translated sentences as next word predictions. 


\section*{Acknowledgments}
The present research was supported by the National Natural Science Foundation of China (Grant No. 61622209). We would like to thank three anonymous reviewers
for their insightful comments.


\bibliography{ref18}
\bibliographystyle{acl}

\end{CJK}
\end{document}